\documentclass[runningheads]{llncs}
\usepackage{graphicx}
\usepackage{comment}
\usepackage{amsmath,amssymb} % define this before the line numbering.
\usepackage{color}
\usepackage{multirow}

\usepackage{mathrsfs}
\usepackage{subfigure}
\usepackage{mathabx}

\makeatletter
\newcommand*\bigcdot{\mathpalette\bigcdot@{.5}}
\newcommand*\bigcdot@[2]{\mathbin{\vcenter{\hbox{\scalebox{#2}{$\m@th#1\bullet$}}}}}
\makeatother

\begin{document}
% \linenumbers
\pagestyle{headings}
\mainmatter
\def\ECCVSubNumber{751}  % Insert your submission number here

\title{Soft Expert Reward Learning for Vision-and-Language Navigation} % Replace with your title

\begin{comment}
\titlerunning{ECCV-20 submission ID \ECCVSubNumber} 
\authorrunning{ECCV-20 submission ID \ECCVSubNumber} 
\author{Anonymous ECCV submission}
\institute{Paper ID \ECCVSubNumber}
\end{comment}
%******************

% \begin{comment}
\titlerunning{Soft Expert Reward Learning for Vision-and-Language Navigation}
\author{Hu Wang \and
Qi Wu \and
Chunhua Shen}
\authorrunning{H. Wang et al.}
\institute{The University of Adelaide, Australia
\email{\{hu.wang,qi.wu01,chunhua.shen\}@adelaide.edu.au}}
% \end{comment}
%******************
\maketitle

\begin{abstract}
Vision-and-Language Navigation (VLN) requires an agent to find a specified spot in an unseen environment by following natural language instructions. Dominant methods based on supervised learning clone expert's behaviours and thus perform better on seen environments, while showing restricted performance on unseen ones. Reinforcement Learning (RL) based models show better generalisation ability but have issues as well, requiring large amount of manual reward engineering is one of which. In this paper, we introduce a Soft Expert Reward Learning (SERL) model to overcome the reward engineering designing and generalisation problems of the VLN task. Our proposed method consists of two complementary components: Soft Expert Distillation (SED) module encourages agents to behave like an expert as much as possible, but in a soft fashion; Self Perceiving (SP) module targets at pushing the agent towards the final destination as fast as possible. Empirically, we evaluate our model on the VLN seen, unseen and test splits and the model outperforms the state-of-the-art methods on most of the evaluation metrics.

\end{abstract}

\section{Introduction}

Vision-and-Language Navigation (VLN) tasks \cite{anderson2018vision} define 
a comprehensive problem: an embodied agent is placed at a spot in a photo-realistic house and the agent is called to navigate to a specific spot based on given natural language instructions. Rising research interests have been put into the VLN since multi-modal data are involved. One of the biggest challenges for this task is to ask an agent to perform appropriate actions in an unseen environment. This in turn requires the agent to learn human behaviours to understand and explore the scene, instead of memorising it.

Current VLN models \cite{anderson2018vision,fried2018speaker,ke2019tactical,ma2019self,ma2019regretful} rely much on behavioural cloning (BC) that treats expert behaviours as strong supervision signals. By doing this, it enables the agents to gain better performance on seen scenarios, however the agents meet trouble on unseen environments due to the error accumulation. As stated in \cite{reddy2019sqil}, teacher forcing models suffer from distribution shift issues because of the greediness of imitating demonstrated expert actions.

Some other works \cite{tan2019learning,wang2019reinforced}, instead, adopt reinforcement learning (RL) along with supervised learning methods intending to overcome the error accumulation issue caused by hard behavioural cloning. However, the reward engineering in RL suffers issues: the reward functions designed at one environment/task may not generalise well to other scenarios; in many practical and complicated tasks, it is hard to define concrete reward functions as game scores. What is more, a hand-crafted reward is defined to target at a certain functionality, it thus inevitably incurs lacking comprehensive considering of the system dynamics. The designing of a reward function requires careful manual tuning and it also suffers generalisation problem due to environment-oriented reward designing, which may affect the model performance while inference.

In this paper, we propose a Soft Expert Reward Learning (SERL) model to address above issues. Our proposed method consists of two orthogonal parts: the Soft Expert Distillation (SED) module that portrays the expert data distribution by distilling knowledge from a random projection space and a Self Perceiving (SP) module that encourages agents to reach the goal as soon as possible.
For the SED module, intuitively, a higher reward should be assigned to an agent who takes an action ``close'' to its expert. To measure the similarity continuously, a density function was adopted to reflect this process in a soft manner rather than leveraging behaviour cloning directly. This density function is implemented to calculate the similarity between observation-action pairs of the expert and the agent in a randomly projected space, by doing which it transforms the expert behaviour into a soft reward signal for the reinforcement learning branch. For the Self Perceiving (SP) module, our model first predicts the schedule to the target location and then utilises the predicted schedule information as an additional reward. As a result, the agent can perceive its current schedule and use it to further pushing itself forward to the goal.

The two newly designed reward modules work complementarily: the Soft Expert Distillation (SED) reward encourages agents to behave as an expert, but the soften behaviour-imitation process makes it more robust; Self Perceiving (SP) module targets at pushing the agents towards the final destination by introducing the current schedule information as another intrinsic reward signal. In summary, this paper makes the following three main contributions.

\begin{itemize}

\item We propose a Soft Expert Distillation (SED) formulation, which is very simple yet offers a highly effective reward signal for obtaining expressive navigational ability. The SED reward encourages the agent to have a better alignment with its expert in a soft manner.
\item We introduce another complementary reward signal with aforementioned SED reward termed as Self Perceiving reward that can help the agent use the current schedule information to push itself to reach the destination as soon as possible.
\item As a result, we show our instantiated model termed as SERL that enables better performance than current state-of-the-art competing methods in both validation unseen and test unseen set of VLN Room-to-Room dataset \cite{anderson2018vision}.
\end{itemize}

\section{Related Work}
\subsection{Vision-and-Language Navigation}

In order to gain promising performance on Vision-and-Language (VLN) \cite{anderson2018vision} task, numerous methods have been proposed, as listed in Table \ref{tab:methods_description}. Many existing works adopt supervised learning and behaviour cloning based methods. Seq2seq \cite{anderson2018vision} model is the most naive baseline that utilises an LSTM-based sequence-to-sequence architecture with attention mechanism to predict the next action. Speaker-Follower \cite{fried2018speaker} model designs a language model (``speaker'') to learn the relationship between visual and language information, as well as a policy network (``follower'') to take actions based on multi-modal inputs. It uses ``speaker'' to synthesise new instructions for data augmentation and help the policy network to select routes. \cite{ke2019tactical} claims its proposed FAST model is able to balance local and global signals while exploring an unobserved environment. It enables the agent act greedily but allows the agent backtrack if necessary according to global signals. \cite{ma2019self} proposes a visual-language co-grounding framework named as self-monitoring model to better fuse the instructions and visual inputs. Building upon self-monitoring model, \cite{ma2019regretful} provides a strategy for the agent to retrieve and re-choose paths based on monitored progress.

Reinforcement learning \cite{mnih2013playing,schulman2017proximal,lillicrap2015continuous} is another paradigm for parameter optimisation. Wang \textit{et al.} \cite{wang2019reinforced} propose a novel Reinforced Cross-modal Matching (RCM) via reinforcement learning to enforce cross-modal matching locally and globally along with imitation learning. In RCM model, an extrinsic reward measuring the reduced distance toward the target location after taking actions, as well as an intrinsic cross-modal matching reward between trajectories and instructions, are proposed. Most recently, \cite{tan2019learning} introduces a novel environment dropout to drop features channel-wisely targeting at feature maps inconsistency issue through combining behaviour cloning and reinforcement learning.

\begin{table}[b]
\caption{Performance Evaluation across different methods.}
\vspace{-10pt}
\label{tab:methods_description}
\begin{center}
\resizebox{0.7\linewidth}{!}{
\begin{tabular}{l|c|c|c|c}
\multirow{2}{*}{Methods}                & Behaviour & Reinforcement & Reward & Reward \\ 
                &  Cloning &  Learning &  Engineering &  Learning \\ \hline
Random \cite{anderson2018vision}                 &                   &                        &                &                 \\
Seq2seq \cite{anderson2018vision}                & \checkmark        &                        &                &                 \\
Speaker-Follower \cite{fried2018speaker}       & \checkmark        &                        &                &                 \\
FAST \cite{ke2019tactical}                   & \checkmark        &                        &                &                 \\
Reinforced Cross-Modal \cite{wang2019reinforced} & \checkmark        & \checkmark             & \checkmark     &                 \\
Self-Monitoring \cite{ma2019self}        & \checkmark        &                        &                &                 \\
Regretful Agent \cite{ma2019regretful}        & \checkmark        &                        &                &                 \\
EnvDrop \cite{tan2019learning}                & \checkmark        & \checkmark             & \checkmark     &                 \\ \hline
SERL (Ours)                 & \checkmark        & \checkmark             & \checkmark     & \checkmark     
\end{tabular}
}\end{center}
\vspace{-15pt}
\end{table}

However, these approaches require either exact imitation of the expert demonstrations or careful reward designing. Behaviour cloning techniques unfortunately lead to error accumulation and further result in catastrophic failure while the agent is exploring unknown environments. Moreover, reward engineering requires careful manual tuning, which motivates us to propose SERL model to learn reward functions from the expert distribution directly.

\subsection{Reward Learning}

Reward engineering is commonly used to design reward functions for reinforcement learning algorithms. In conventional reinforcement learning tasks, such as playing Atari games \cite{brockman2016openai}, rewards are individually shaped by each game simulators. However, reward engineering has obvious drawbacks --- the reward functions are designed targeting at different environments which is not generic. There are some methods have been proposed to solve this problem. Recently, Inverse reinforcement learning (IRL) \cite{ng2000algorithms} framework is proposed to extract reward functions from expert behaviours by updating both of the reward functions and the policy networks. Random Expert Distillation (RED) \cite{wang2019random} proposed an expert policy support estimation method to distil rewards from given expert trajectories. Generative Adversarial Imitation Learning (GAIL) \cite{ho2016generative} is also a recently proposed model which tries to bypass the reward function and learn experts behaviour directly with generative adversarial networks.

Comparing with the IRL and GAIL models, our proposed Soft Expert Distillation module learns expert demonstration data distribution directly by comparing the output similarity between a randomised network and a distillation network, rather than utilising iterative model updating and generative adversarial networks. The RED model designs state and action in relatively small spaces for the Mujoco environment \cite{todorov2012mujoco} and its driving task; while we design our SED module in fundamentally different state and action spaces for navigation in photo-realistic Matterport3D environments. We are the first to introduce soft expert reward learning framework into Vision-and-Language task.

\section{Soft Expert Reward Learning Model}
\subsection{Overview and Problem Definition}

\begin{figure}[t]
\begin{center}
% \hspace*{-0.2cm}
\includegraphics[width=0.85\textwidth]{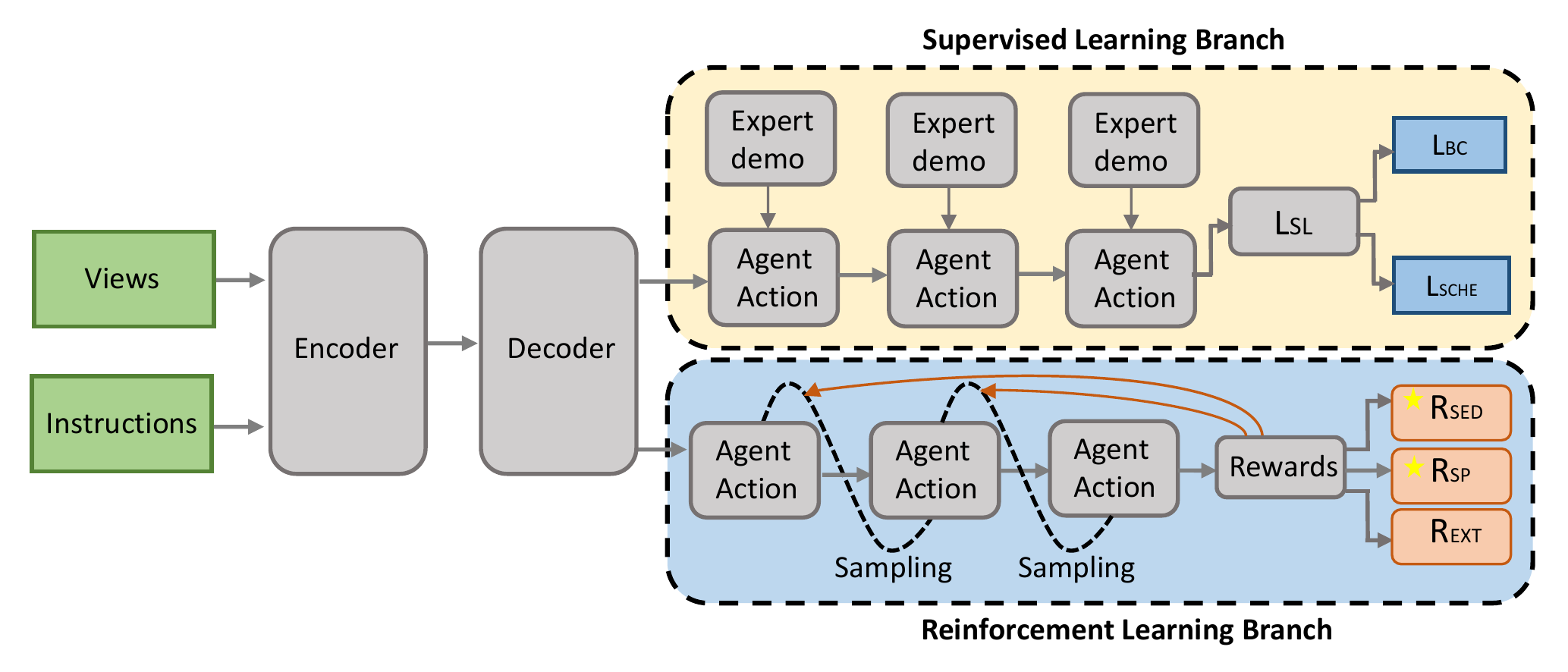}
\end{center}
\vspace{-0.6cm}
\caption{The proposed Soft Expert Reward Learning (SERL) framework. After getting the visual features and language features through the encoder, they are fed into the decoder to obtain the selected action $a_t$ for time step $t$. The training process of SERL is divided into two parts: a supervised learning branch and a reinforcement learning branch. We introduce two novel rewards (marked with yellow stars in the figure): Soft Expert Distillation (SED) reward and Self Perceiving (SP) reward.}\label{fig:framework}
\end{figure}

Vision-and-Language Navigation task requires an agent placed at a unknown photo-realistic house to understand multi-modal data comprehensively, so that the agent can navigate to the specified location. The multi-modal data includes natural image data and natural language instructions. More specifically, after an agent is spawn, at each time step $t$ the observation of the agent consists of 36 images of panoramic views, denoted as $V_t = \{v_{t,1}, v_{t,2}, ..., v_{t,36}\}$. The navigable views $N_t = \{n_{t,1}, n_{t,2}, ..., n_{t,k}, n_{t,k+1}\}$ are given as well, where $k$ denotes the maximum number of navigable viewpoints and $n_{t,k+1}$ represents ``stay'' action. A $m$ words length instruction is given which is denoted as $X = \{x_1, x_2, ..., x_m\}$. Based on the visual and language information, actions at each time $a_t$ will be selected and eventually a trajectory $\tau = \{a_1, a_2, ..., a_T\}$ is formed. The objective of VLN task is to find the optimal action $a^*_t$ at each step to quickly reach the target location, while keep the trajectory $\tau$ as short as possible. Since Vision-and-Language Navigation task is a sequential decision problem, it can be modelled as a Markov Decision Process (MDP), which is noted as a four-element-tuple ($\mathcal{S}, \mathcal{A}, \mathcal{P}, \mathcal{R}$). $\mathcal{S}$ and $\mathcal{A}$ represent state and action sets relatively. $\mathcal{P}$ is the environment dynamics and it can be presented in the form $\mathcal{P}(s, s') = P(s'|s, a)$. $\mathcal{R}$ is the reward function.

In this paper we introduce a Soft Expert Reward Learning model to distil reward function directly from expert demonstrations and soften the process of behaviour cloning to alleviate the drawbacks from error accumulation. The structure of our model is illustrated through Figure \ref{fig:framework}. We follow a standard Encoder-Decoder paradigm. The encoder plays the role as a multi-modal data feature extractor to fetch the features from both visual images and language instructions. The decoder is a LSTM (long short-term memory) network with attention mechanism to predict actions according to the abovementioned two branches: the supervised learning branch helps the agent imitate the expert demonstration and perceive the current schedule to the target location; the reinforcement learning branch optimises the outputted action probability distribution from reinforcement learning aspects. The key difference of our proposed SERL model with previous models is that we proposed two novel intrinsic reward signals: Soft Expert Distillation reward $R_{SED}$ encourages the agent to align with expert actions but in a soft fashion and Self Perceiving reward $R_{SP}$ motivates the agent to reach the goal as fast as possible with predicted schedule information. In the following sections, we will first introduce the Encoder-Decoder structure and then introduce the two reward functions.

\subsection{Encoder-Decoder Structure}

\begin{figure}[t]
\begin{center}
% \hspace*{-0.2cm}
\includegraphics[width=1.0\textwidth]{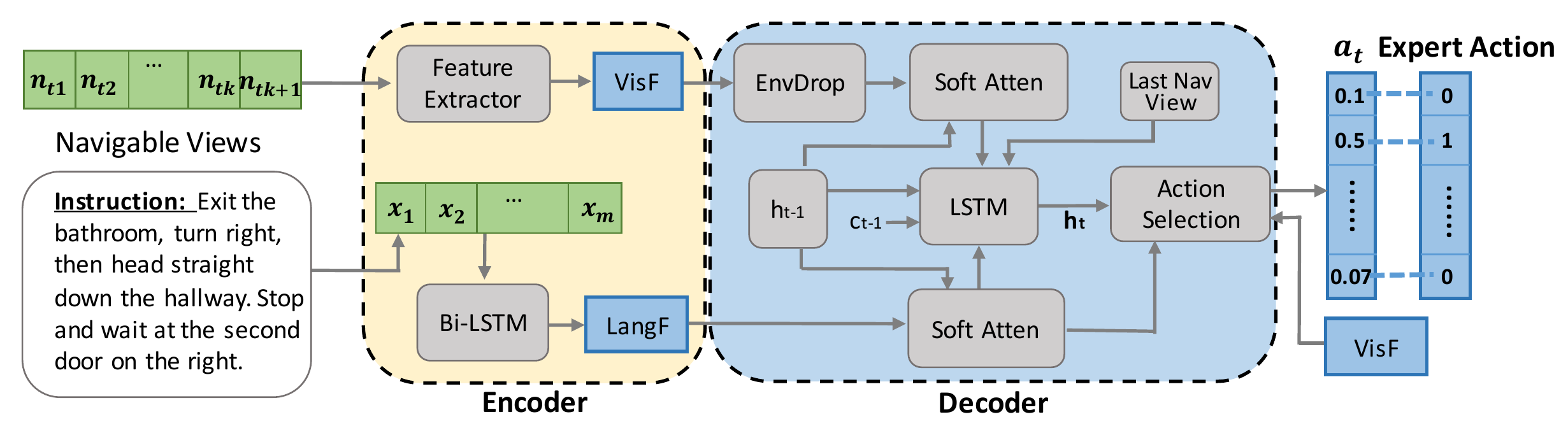}
\end{center}
\vspace{-0.5cm}
\caption{Encoder-Decoder Structure of Soft Expert Reward Learning (SERL) framework. After fetching the visual and language features from the encoder, the multi-modal features are fed into the decoder to obtain cross-modal attentions. Finally, actions will be chosen according to the attentive features.}\label{fig:enc-dec}
\end{figure}

Encoder-Decoder structure (as shown in \ref{fig:enc-dec}) is adopted as the main structure of our method. Natural image data and natural language instructions are inputted to an encoder to extract corresponding features maps. Following the paper \cite{ma2019self,tan2019learning}, we extract ResNet \cite{he2016deep} features of the navigable views concatenated with the orientation as the visual features $VisF_t$. We then use a Bi-Directional Long Short-Term Memory (Bi-LSTM) to pull out language features $LangF_t$. The multi-modal features are fed into a decoder to output the next action probability vectors later on.

\paragraph{Encoder:} On the encoder side, after pre-extracting ResNet features of different views, the feature maps of each navigable view $n_{t,i}$ is attached with an orientation tag $(\cos\gamma_{t,i},\sin\gamma_{t,i},\cos\varphi_{t,i},\sin\varphi_{t,i})$ to form the visual feature $VisF_t$:
\begin{equation}
    VisF_t = concat(resnet(n_{t,i}), (\cos\gamma_{t,i},\sin\gamma_{t,i},\cos\varphi_{t,i},\sin\varphi_{t,i})),
\end{equation}
where $concat(.)$ is a concatenation function.

For the language perspective, after each word of the instruction is tokenised into a vector, the token vectors are fed into a Bi-LSTM network to extract the language features $LangF_t$. As Eqn. \ref{eqn:langf}, formally we have
\begin{equation}\label{eqn:langf}
    LangF_t = \{x'_{1}, x'_{2}, ..., x'_{m}\} = Bi\mbox{-}LSTM(\{x_1, x_2, ..., x_m\}),
\end{equation}
where $x'_{i}$ is the corresponding i-th encoded word tokenised by Bi-LSTM.

\paragraph{Decoder:} On the decoder side, after the visual feature $VisF_t$ and language features $LangF_t$ are formed, along with the last cross-modal hidden state $h_{t-1}$, they are fed into soft attention layers to fetch the attentive visual and language features. Following the work \cite{tan2019learning}, the environment dropout is used on $VisF_t$ before feeding into soft attention layer to obtain feature-wise dropout for consistency in different views. Formally,
\begin{equation}
    V\widetilde{isF}_t = Soft\mbox{-}Atten(EnvDrop(VisF_t), h_{t-1}),
\end{equation}
\begin{equation}
    L\widetilde{ang}F_t = Soft\mbox{-}Atten(LangF_t, h_{t-1}).
\end{equation}
Together with previous navigated view $pre^v_{t-1}$, last cross-modal hidden state $h_{t-1}$, cell state $c_{t-1}$, attentive visual and language features are fed into a LSTM layer to form the cross-modal hidden state $h_t$ and cell state $c_t$ at step $t$. This step is critical for the model to fuse the visual and language multi-modal signals to choose the action.
\begin{equation}
    h_t, c_t = LSTM(h_{t-1}, c_{t-1}, pre^v_{t-1}, V\widetilde{isF}_t, L\widetilde{ang}F_t).
\end{equation}
The action probability distribution for the next step is calculated as:
\begin{equation}
    p_t = softmax(fc(L\widetilde{ang}F_t, drop(h_t)) \cdot VisF_t),
\end{equation}
where $drop(.)$ represents a dropout function. The dot product $\cdot$ is used hereafter for matrix multiplication operation.

The decoder is connected to two branches: supervised learning branch and reinforcement learning branch. These two branches optimise the outputted action probability distribution from two different learning paradigms. In this case, the total loss function is:
\begin{equation}
    L = L_{SL} + L_{RL}.
\end{equation}

\paragraph{SL Branch:} In the supervised learning branch, the cross-entropy loss between the predicted action logits and expert actions one-hot vector is calculated to force the agent to mimic its teacher's behaviours. This loss is termed as behaviour cloning loss $L_{BC}$. Following the work \cite{ma2019self}, besides the behaviour cloning loss, another loss to predict current schedule towards the goal is adopted. This loss is named as schedule loss $L_{SCHE}$ working as an additional supervisory signal. Formally, the loss function for the supervised learning branch is:
\begin{equation}
    L_{SL} = L_{BC} + L_{SCHE}.
\end{equation}
where the behaviour cloning loss $L_{BC}$ can be presented detailedly:
\begin{equation}
    L_{BC} = - \sum_i^n y_{t,i}^{act} log(p_{t,i}),
\end{equation}
where $p_{t}$ and $y_{t}^{act}$ are predicted action logits and expert actions one-hot vector at step $t$ respectively. 

To calculate the $L_{SCHE}$, the model ought to predict distance improvement ratio in advance at each step as its current schedule information. Then, L2 distance between predicted schedule and the genuine schedule is chosen as the loss function. Formally,
\begin{equation}
    L_{SCHE} = (y_{t}^{sche} - V_t^{sche})^2,
\end{equation}
where $V_t^{sche}$ represents the predicted schedule which will be described in detail in the subsequent section and $y_{t}^{sche}$ is the corresponding true schedule value.

\paragraph{RL Branch:} As the reinforcement learning branch shown in Figure \ref{fig:framework}, we adopt actor-critic algorithm \cite{mnih2016asynchronous} as our reinforcement learning method. For the reinforcement learning branch, the training loss $L_{RL}$ can be formally represented as:
\begin{equation}
    L_{RL} = \underbrace{\sum_t -log(p_t) * (\widebar{R_t} - v(h_t))}_{actor\ loss} + \underbrace{\sum_t (\widebar{R_t} - v(h_t))^2}_{critic\ loss},
\end{equation}
where $v(.)$ is the value function of critic. $\widebar{R_t}$ represents the discounted reward for time step $t$ and it can be formulated as:
\begin{equation}
    \widebar{R_t} = \widebar{R_{t+1}} * \gamma + R_t,
\end{equation}
in which the $\gamma$ is the discount factor. The reward $R_t$ is made up of three parts: an extrinsic reward $R_{EXT}$ and another two complementary and newly proposed reward functions --- Soft Expert Distillation (SED) reward $R_{SED}$ and Self Perceiving reward $R_{SP}$. The total reward function thus can be formalised as:
\begin{equation}
    R_t = \alpha R_{SED} + \beta R_{SP} + R_{EXT},
\end{equation}
where (1) SED reward $R_{SED}$, an automatically learnt reward function through aligning agent's behaviours to the provided expert demonstrations. (2) SP reward $R_{SP}$, a reward function comes from predicted schedule to encourage the agent to reach the goal as soon as possible. (3) The extrinsic reward $R_{EXT}$ assigns the agent a positive reward, if the agent stops within three-meter from target or the agent reduces the distance to the goal; otherwise, a negative reward will be returned. $\alpha$, $\beta$ are the trade-off factors of SED reward and SP reward respectively. The details of individual proposed reward function will be revealed in the following sections.

\subsection{Soft Expert Distillation}

\begin{figure}[t]
\begin{center}
% \hspace*{-0.2cm}
\includegraphics[width=0.8\textwidth]{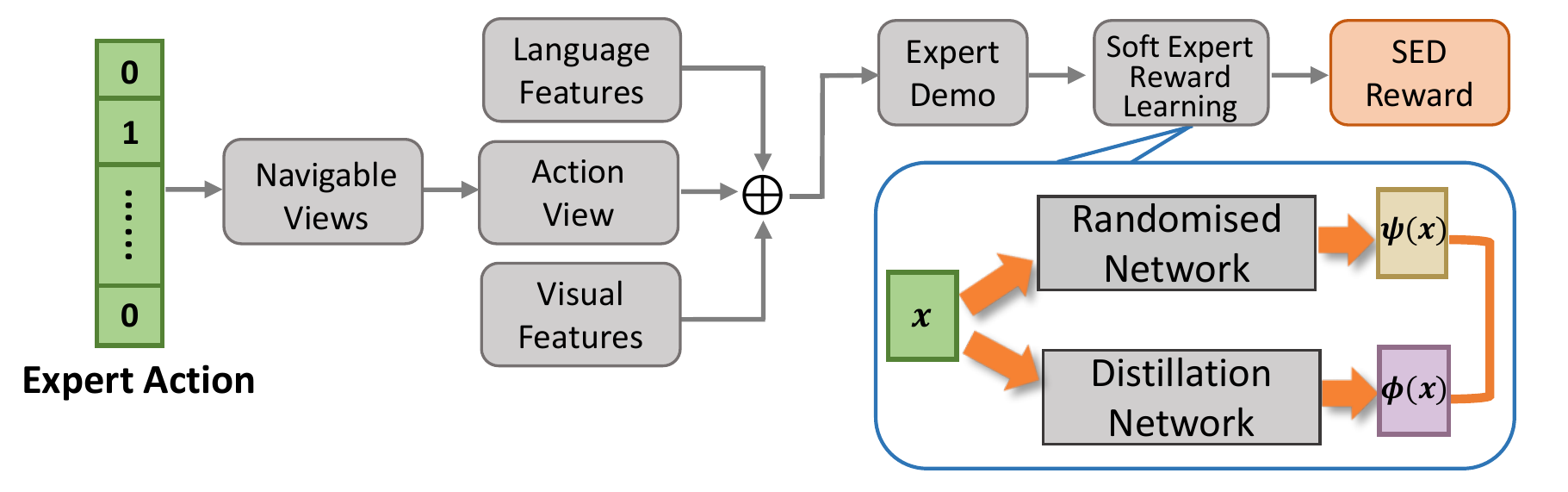}
\end{center}
\vspace{-0.5cm}
\caption{The Soft Expert Distillation networks structure. Given an expert demonstrated data point $x \in \mathbb{R}^{N}$, it is fed into a weight-fixed randomly initialised neural network $\psi(\mathbf{x})$; simultaneously, the data point $x$ is inputted into a distillation network $\phi(\mathbf{x};\theta)$ with different structure but same output dimensions with the parameters $\theta$.}\label{fig:sed}
\end{figure}

Inspired by the work \cite{wang2019random}, we propose to learn the reward function from inputted expert demonstration in Vision-and-Language Navigation task. We train a neural network to predict the output of a random-initialised but frozen network to distil the expert knowledge. The Soft Expert Distillation networks structure is shown in Figure \ref{fig:sed}. The key intuition behind this is: given a certain amount of random projection information, the representation learner is required to fit the structure of these given data points in the random projection space to achieve a similar projected distribution. The learning function is expected to predict relatively better where more expert data lays. In this case, a strong density function is formed. It models the likelihood of the agent performing a similar action with its expert in a situation through distillation. A higher prediction distance, which results in a low SED reward in turn, will be assigned to unexpected observation-action pairs that differs from given expert demonstrations. Thus, a higher reward will be assigned to an agent who takes an action similar with its expert. This encapsulation of density function gives us another view of learning expert demonstrations directly other than \cite{ng2000algorithms} and \cite{ho2016generative}.

Precisely, for a given expert demonstrated data point $x \in \mathbb{R}^{N}$, we ﬁrst feed it into a weight-fixed and random-initialised neural network $\psi(\mathbf{x})$; at the same time the data point $x$ is inputted into a distillation network $\phi(\mathbf{x};\theta)$ with different structure but same output dimensions. The data is projected to a $M$-dimensional new space by a representation learner $\phi:\mathbb{R}^{N} \mapsto \mathbb{R}^{M}$ with the parameters $\theta$. We emphasise here, the function capacity of network $\phi$ is less than network $\psi$, by doing which can prevent overfitting. As we adopt L2 distance as our loss function, then we formulate the subsequent step as a prediction task and define a loss function as:

\begin{equation}
    L_{sed}(\mathbf{x})=\left(\phi(\mathbf{x};\theta), \psi(\mathbf{x}) \right)^{2},
\end{equation}

Empirically, both of $\psi$ and $\phi$ are implemented by multi-layer perceptrons. $\psi:\mathbb{R}^{N}\mapsto\mathbb{R}^{M}$ plays the role of a random data mapping function to project points into a randomly projected space. By doing so, this loss offers a simple yet powerful supervisory signal for the distillation network to learn semantic-rich feature representations from given expert data processed by the random projection function $\psi$.

In order to distil the expert behaviour distribution, the data points are consist of expert's visual observation, language instructions and actions. The equation is formally shown as:

\begin{equation}
    L^{t}_{sed}=\left(\phi(\{VisF_t, LangF_t, a_t\};\theta) - \psi(\{VisF_t, LangF_t, a_t\}) \right)^{2}.
\end{equation}

The SED module preserves semantic-rich information w.r.t. distribution of expert demonstration for the representation learner. So the module is an ideal density function to measure the similarity of an agent's behaviour with the expert demonstration. Differ from the behaviour cloning process, it is formed in a soft manner. The SED intrinsic reward function is formally presented as:

\begin{equation}
R_{SED}= 
\left\{
             \begin{array}{lr}
             +2, if Dis^{sed}_t <= thresh \\
             -2, if Dis^{sed}_t > thresh
             \end{array}
\right.
\end{equation}

The L2 distance between $\phi(\{VisF_t, LangF_t, a_t\};\theta)$ and $\psi(\{VisF_t, LangF_t, a_t\})$ is denoted as $Dis^{sed}_t$. Intuitively, if $Dis^{sed}_t$ is less than the threshold, it represents the current behaviour of the agent is similar with the expert distribution where a positive reward should be awarded; otherwise, a negative reward will be returned. In contrast, behaviour cloning based models encourage the agent to copy expert demonstrations exactly; while our proposed soft expert distillation module learns the demonstrated behaviour in a soft manner by depicting the distribution of expert behaviours. In the case, the agent can retain the expert knowledge but will not suffer from the error accumulation problem. Thus, it increases the robustness of the model across various VLN environments.

\subsection{Self Perceiving Reward}

To perceive the schedule information towards the goal is crucial for the agent to complete the VLN task. A self perceiving module is designed to predict distance improvement ratio at each step as current schedule information of the agent. In order to utilise the information more adequately, we take one more step ahead by making use of this schedule information as another intrinsic reward---self perceiving reward. Formally, the self perceiving reward is calculated from:

\begin{equation}
    C_{attn} = softmax(fc(h_{t-1}) \cdot LangF_t),
\end{equation}

\begin{equation}
    R_{SP} = V_t^{sche} = \sigma(fc(drop(\tanh (c_t) \odot \sigma(fc(h_{t-1}, V\widetilde{isF}_t))), C_{attn})),
\end{equation}

where $C_{attn}$ represents the language attention over different vocabularies within the instruction sentence. $\odot$ is the element-wise Hadamard product. Intuitively, the Self Perceiving reward indicates the predicted schedule information toward the destination. The more distance improvement ratio of the current action archived, the higher reward ought to be assigned. Moreover, this reward offers more information of distance change than raw distances. The more self perceiving reward the agent collected, the closer the agent believes to reach the target location.

\section{Experiments}

Following previous works \cite{anderson2018vision,fried2018speaker,ke2019tactical,ma2019self,ma2019regretful,tan2019learning,wang2019reinforced}, we evaluate our model on the Room-to-Room (R2R) dataset \cite{anderson2018vision} for VLN task. Furthermore, we test our method on the VLN test server\footnote{The VLN leaderboard address is https://evalai.cloudcv.org/web/challenges/challenge-page/97/leaderboard/270.} \cite{EvalAI} to validate the proposed Soft Expert Reward Learning Model. Ablation study is further conveyed to examine the contribution of each individual component of the model. The experimental results show the effectiveness of the proposed model.

\subsection{Experimental Setup}

\noindent\textbf{Evaluation Metrics.} Currently, a variety of metrics are used to evaluate VLN models. We adopt the following metrics: Navigation Error (NE) is to measure the shortest path distance between the stopping position and the goal; Success Rate (SR) quantifies the rate of success if the agent can stop within three meters from the target; Oracle Success Rate (OSR) is the success percentage if the agent can stop at the closest point along its trajectory; the Success rate weighted by Path Length (SPL) \cite{anderson2018evaluation} is also adopted to indicate the weighted SR.

\noindent\textbf{Implementation Detail.} Following \cite{fried2018speaker,tan2019learning}, we utilise the ResNet-152 model pre-trained on ImageNet to extract CNN features as visual inputs. Empirically, we set the $M$ equal to 128, and set both of the reward trade-off factors $\alpha$ and $\beta$ to 0.1. In Soft Expert Distillation networks, the randomised network is made up of two hidden linear layers with 512 and 256 neurons respectively; the distillation network has one hidden linear layers with 256 neurons. Between every two linear layers, both of the randomised network and the distillation network adopt leaky-relu as their activation function. To prevent overfitting, we early-stopped the training process of models according to the performance on the validation set. The Soft Expert Distillation module is not jointly trained with the rest of the model. This decoupling prevents performance unstableness during training and increase the robustness of the model.

\subsection{Overall Performance}

\begin{table}[t]
\caption{Performance Evaluation across different methods. The first place of each column is bolded. All of the results are reported on models without beam search, except FAST \cite{ke2019tactical} model using a beam-search style strategy. The $\uparrow$ means that the higher the better; vice versa. The * sign represents data augmentation.}
\label{tab:performance}
\begin{center}
\hspace*{-0.4cm}
% \resizebox{\textwidth}{23mm}{
\scalebox{0.8}{
\setlength{\tabcolsep}{1mm}{
\begin{tabular}{l|c c c c|c c c c|c c c c}
                       & \multicolumn{4}{c}{Val Seen} & \multicolumn{4}{c}{Val Unseen} & \multicolumn{4}{c}{Test Unseen} \\
Methods                 & NE $\downarrow$            & SR $\uparrow$            & OSR $\uparrow$           & SPL $\uparrow$           & NE $\downarrow$            & SR $\uparrow$            & OSR $\uparrow$           & SPL $\uparrow$           & NE $\downarrow$            & SR $\uparrow$            & OSR $\uparrow$           & SPL $\uparrow$           \\ \hline
Random \cite{anderson2018vision}                  & 9.45          & 0.16          & 0.21          & -             & 9.23          & 0.16          & 0.22          & -             & 9.77          & 0.13          & 0.18          & -             \\
Seq2seq \cite{anderson2018vision}                 & 6.01          & 0.39          & 0.53          & -             & 7.81          & 0.22          & 0.28          & -             & 7.85          & 0.20           & 0.27          & 0.18          \\
Self-Monitoring \cite{ma2019self}         & 3.72          & 0.63          & \textbf{0.75} & 0.56          & 5.98          & 0.44          & 0.58          & 0.30           & -             & -             & -             & -             \\
Regretful-Agent \cite{ma2019regretful}         & 3.69          & 0.65          & 0.72          & \textbf{0.59} & 5.36          & 0.48          & \textbf{0.61}          & 0.37          & -             & -             & -             & -             \\
EnvDrop \cite{tan2019learning}                 & 4.71          & 0.55          & -             & 0.53          & 5.49          & 0.47          & -             & 0.43          & -             & -             & -             & -             \\
SERL (Ours)             & \textbf{3.67} & \textbf{0.66} & 0.71          & 0.58          & \textbf{4.97} & \textbf{0.50}  & 0.59 & \textbf{0.44} & \textbf{5.70}  & \textbf{0.51} & \textbf{0.57} & \textbf{0.47} \\ \hline
Speaker-Follower* \cite{fried2018speaker}       & 3.36          & 0.66          & 0.74          & -             & 6.62          & 0.36          & 0.45          & -             & 6.62          & 0.35          & -             & 0.28          \\
RCM* \cite{wang2019reinforced} & 3.37          & 0.67          & 0.77          & -             & 5.88          & 0.43          & 0.52          & -             & 6.12          & 0.43          & 0.50           & 0.38          \\
FAST* \cite{ke2019tactical}                   & -             & -             & -             & -             & 4.97          & \textbf{0.56}          & -             & 0.43          & \textbf{5.14} & \textbf{0.54} & -             & 0.41          \\
Self-Monitoring* \cite{ma2019self}        & 3.22          & 0.67          & \textbf{0.78} & 0.58          & 5.52          & 0.45          & 0.56          & 0.32          & 5.67          & 0.48          & 0.59          & 0.35          \\
Regretful-Agent* \cite{ma2019regretful}       & 3.23          & \textbf{0.69}          & 0.77          & 0.63          & 5.32          & 0.50           & 0.59          & 0.41          & 5.69          & 0.48          & 0.56          & 0.40           \\
EnvDrop* \cite{tan2019learning}               & 3.99          & 0.62          & -             & 0.59          & 5.22          & 0.52          & -             & \textbf{0.48}          & 5.23          & 0.51          & 0.59          & 0.47          \\
EnvDrop-Our-Impl*                & 3.77            & 0.66          & 0.72           & 0.62           & 5.49            & 0.49          & 0.56           & 0.45          & -          & -          & -          & -          \\
SERL* (Ours)            & \textbf{3.20}  & \textbf{0.69} & 0.75          & \textbf{0.64} & \textbf{4.74} & \textbf{0.56} & \textbf{0.65} & \textbf{0.48} & 5.63          & 0.53          & \textbf{0.61} & \textbf{0.49}
\end{tabular}
}}
\end{center}
\end{table}

In this section, we convey the evaluation experiments on three individual sets, validation seen, validation unseen and test set, shown in table \ref{tab:performance}, to compare the effectiveness of our proposed soft expert reward learning model with other models. The comparison is split into two groups: models trained on non-augmented data and augmented data. Within twelve indicators of validation set and test set, we achieve ten best results on the non-augmented group and nine best results on the augmented group, which reveals the effectiveness of SERL model. More specifically, for the non-augmented group, on validation unseen set, our SERL model reduces the navigation error by 7\%, increase the success rate by 4\% and SPL by 2\%. Our method also receives remarkable results on test unseen set. Similarly for the augmented group, on validation unseen set, it is clear that our model is the best performer. SERL model reduces the navigation error by 5\% and gets 0.56 successful rate. Our model also increases 10\% for the oracle successful rate and gets 0.48 SPL respectively compared to the second-best model. On the test unseen set, our SERL model can achieve performance better than, or comparably well to, the other competing methods in Table \ref{tab:performance}. When compared to the second-best model, the model increases 3\% for the oracle successful rate and 4\% SPL respectively. The FAST \cite{ke2019tactical} model applies a beam-search style strategy, thus it is expected to produce better successful rate (SR) but it leads to a relatively worse SPL.

\subsection{Ablation Study}
\subsubsection{Ablation Study of Different Components Performance}

This section examines the contribution of each component of SERL model. Different components are added to the baseline model. The ablation results are represented as Table \ref{tab:ablation}. The results are shown on validation seen and unseen sets and the models are trained with the same data augmentation strategy. In the first column, SED represents our proposed soft expert distillation module, while SP is the self perceiving module. BS represents beam search setting. We check different components in the second column to examine each variant. Row model \#1 shows the performance of the environment dropout methods that we implemented.
From the table we can clearly find that when comparing to row \#1, excluding the beam search setting on the validation unseen set, the model with SED module alone (method \#2) achieves higher SR by 6\% and increases SPL score from 0.45 to 0.48; the model with SP module alone (\#3) receives better success rate as 0.53 from 0.49 and better SPL score as 0.46 from 0.45. This is because the SED module encourages the agent to have better alignment with expert trajectories, but in a soft way; the SP module pushes the agent to find the target location as fast as possible. The full SERL model (method \#4) combines the advantages of individual module and it achieves 0.56 of successful rate and 0.48 of SPL, which outperforms other variants.

\begin{table}[t]
\caption{Ablation study of different components in SERL model. We evaluate the results on validation seen set and validation unseen set. The best result are bolded.}
\label{tab:ablation}
\begin{center}
\hspace*{-0.2cm}
% \resizebox{\textwidth}{23mm}{
\scalebox{0.9}{
\setlength{\tabcolsep}{1mm}{
\begin{tabular}{l|ccc|cccc|cccc}
       & \multicolumn{1}{c}{} &            &            & \multicolumn{4}{c|}{Val Seen}                                     & \multicolumn{4}{c}{Val Unseen}                                   \\
Models & SED                   & SP         & BS         & NE $\downarrow$ & SR $\uparrow$ & OSR $\uparrow$ & SPL $\uparrow$ & NE $\downarrow$ & SR $\uparrow$ & OSR $\uparrow$ & SPL $\uparrow$ \\ \hline
1      &                       &            &            & 3.77            & 0.66          & 0.72           & 0.62           & 5.49            & 0.49          & 0.56           & 0.45           \\
2      & \checkmark            &            &            & 3.67            & 0.66          & 0.74           & 0.63           & 5.10            & 0.52          & 0.58           & \textbf{0.48}           \\
3      &                       & \checkmark &            & 3.19            & 0.67          & 0.72           & 0.61           & 4.93            & 0.53          & 0.61           & 0.46           \\
4      & \checkmark            & \checkmark &            & 3.20            & 0.69          & 0.75           & \textbf{0.64}  & 4.74            & 0.56          & 0.65           & \textbf{0.48}  \\
5      & \checkmark            & \checkmark & \checkmark & \textbf{2.47}   & \textbf{0.77} & \textbf{0.99}  & 0.02           & \textbf{3.01}   & \textbf{0.71} & \textbf{0.99}  & 0.02          
\end{tabular}
}}
\end{center}
\end{table}
Additionally, beam search is another popular Vision-and-Language Navigation setting. In the beam search setting, the agents are given the chance to choose the trajectories with the highest success rate. In this case, it can further boost the success rate of our SERL model (method \#5) to 0.77 on validation seen set and 0.71 on validation seen set. Moreover, SERL model receives 0.70 in successful rate on the test unseen set with beam search.

\subsubsection{Sensitivity Test}

\begin{figure}[!t]
\vspace{-0.5cm}
\centering
% \hspace{-0.5cm}
\subfigure{
\begin{minipage}[t]{0.4\linewidth}
\centering
\includegraphics[width=1.05\textwidth]{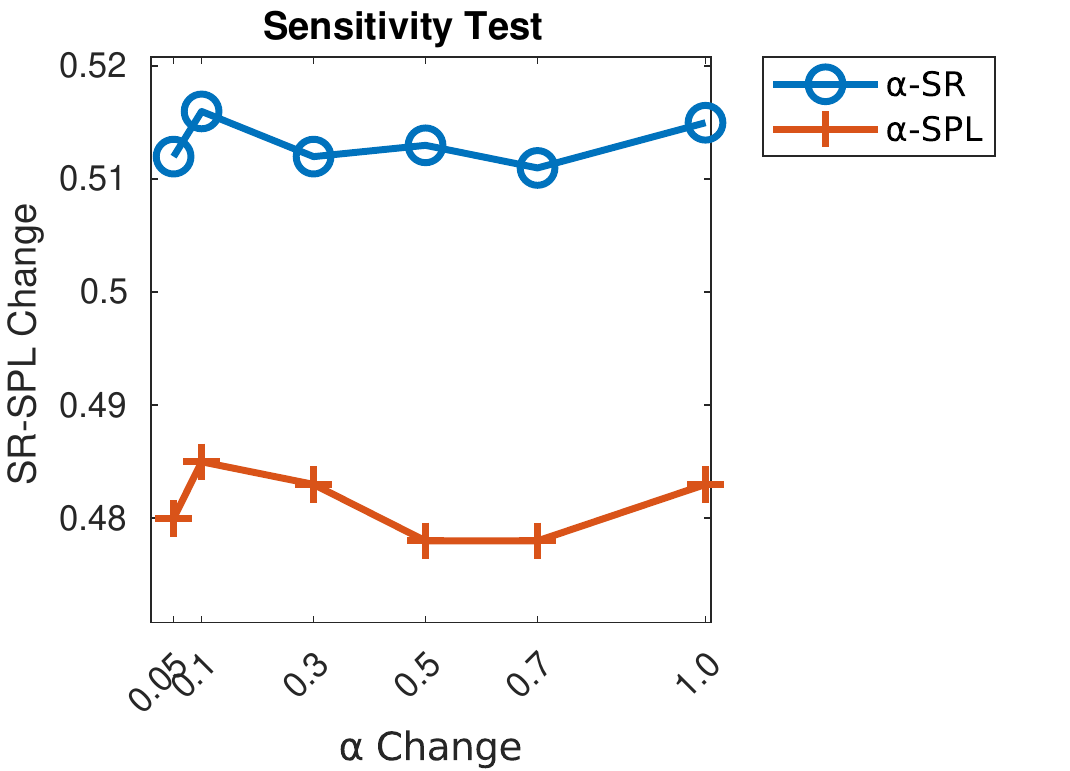}
% \caption{(a)}
\end{minipage}%
}%
\subfigure{
\begin{minipage}[t]{0.4\linewidth}
\centering
\includegraphics[width=1.05\textwidth]{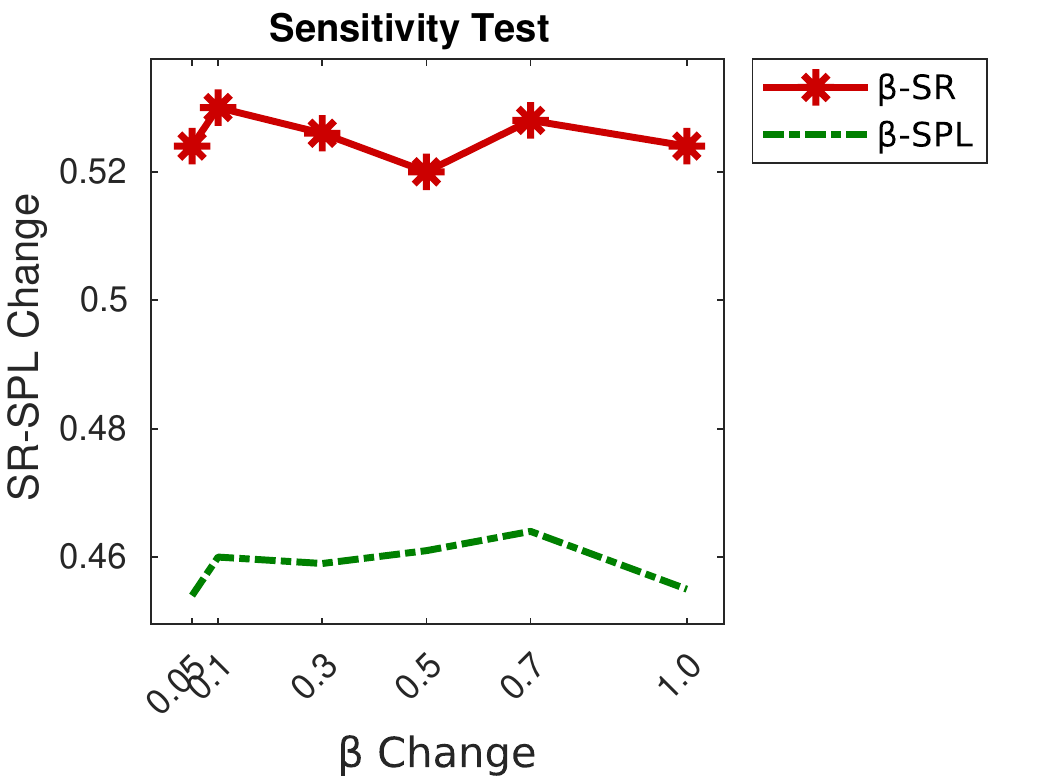}
% \caption{(b)}
\end{minipage}%
}%
\centering
\caption{The sensitivity test of our Soft Expert Reward Learning (SERL) model. The figures show the SR and SPL performance of the model on validation unseen set with different $\alpha$ and $\beta$ values.}\label{fig:sen}
\end{figure}

This section presents the performances of SERL model with different $\alpha$ and $\beta$ weights to trade-off the proposed individual intrinsic reward. Figure \ref{fig:sen} shows the sensitivity test results, which is evaluated in SR and SPL on validation unseen set. It is clear that SERL generally performs stably w.r.t. the use of different $\alpha$ and $\beta$ weights. This demonstrates the general stability of our SERL method by setting different hyper-parameters. In general, $\alpha=\beta=0.1$ is recommended for SERL to achieve effective visual and language navigation performance.

\subsection{Visualisation}

\begin{figure}[t]
\begin{center}
% \hspace*{-0.2cm}
\includegraphics[width=1.0\textwidth]{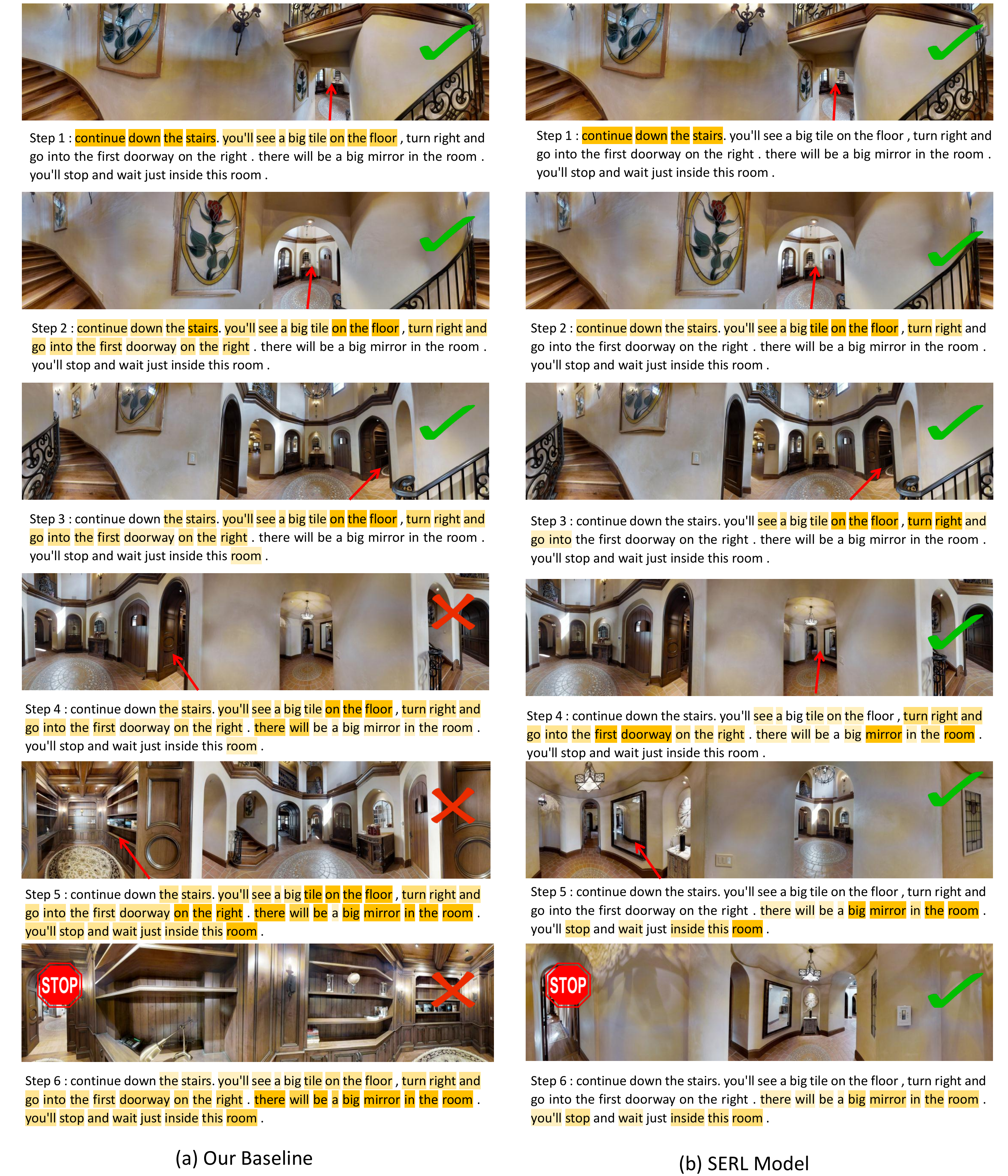}
\end{center}
\caption{The visualisation of our proposed Soft Expert Reward Learning (SERL) model. The figure shows the comparison between SERL model and the baseline model. The yellow colours in the sentence represents the attention maps over the instruction. The depth of the colours indicates the strength of the attention. The darker the colours, the more attention is put on the specific vocabularies. The check mark means the agents take a same action as the expert; the cross mark represents the opposite.}\label{fig:vis}
\end{figure}

Figure \ref{fig:vis} shows the actions taken by our baseline agents and proposed SERL agent, respectively. The attention maps over the instruction at each step are also illustrated in the figure. On the left column of the figure, the agent is trained by behaviour cloning solely and it performs correctly at the first three steps. But the agent takes a wrong action at the fourth step and it results in failure navigation in the next three steps. This is because subtle errors will be accumulated at each step by just copy expert demonstrations in the training phase. However, our SERL model can attend over the instruction in a better way and it does not encounter the error accumulation problem in the case.

\section{Conclusions}

In this paper, we propose a Soft Expert Reward Learning (SERL) model to address the behaviour cloning error accumulation and the reinforcement learning reward engineering issues for VLN task. From the experimental results, we show that our SERL model gains better performance generally than current state-of-the-art methods in both validation unseen and test unseen set on VLN Room-to-Room dataset. The ablation study shows that our proposed the Soft Expert Distillation (SED) module and the Self Perceiving (SP) module are complementary to each other. Moreover, the visualisation experiments further verify the SERL model can overcome the error accumulation problem. In the future, we will further investigate more reward learning methods on VLN task.

\clearpage
\bibliographystyle{splncs04}
\bibliography{mybib}
\end{document}